\documentclass{article}

\pdfoutput=1

\usepackage{natbib}
\PassOptionsToPackage{numbers, compress}{natbib}

\usepackage[preprint]{neurips_data_2023}

\usepackage[utf8]{inputenc} 
\usepackage[T1]{fontenc}    
\usepackage{hyperref}       
\usepackage{url}  
\usepackage{caption} 
\usepackage{booktabs}       
\usepackage{amsfonts}       
\usepackage{nicefrac}       
\usepackage{microtype}      
\usepackage{xcolor}         
\usepackage{soul}
\usepackage{graphicx}
\usepackage{multirow}
\usepackage{ifthen}
\usepackage{float}
\usepackage{longtable}
\usepackage{arydshln}
\newboolean{isBlue}
\setboolean{isBlue}{false} 
\newcommand{\pr}[1]{%
    \ifthenelse{\boolean{isBlue}}%
    {\textcolor{blue}{#1}}
    {#1}
}

\title{Towards a Unified Framework for Adaptable Problematic Content Detection via Continual Learning}

\author{%
  Ali Omrani*\\
  University of Southern Califronia\\
  \texttt{aomrani@usc.edu} \\
  \And  
  Alireza S. Ziabari*\\
  University of Southern California \\
  \texttt{salkhord@usc.edu} 
  \And 
  Preni Golazizian\\
  University of Southern California \\
  \texttt{golazizi@usc.edu} \\
 \And
  Jeffrey Sorensen\\
  Jigsaw \\
  \texttt{sorenj@google.com} \\
\And
  Morteza Dehghani\\
  University of Southern California \\
  \texttt{mdehghan@usc.edu} \\
}
  
\begin{document}

\maketitle

\begin{abstract}
   
Detecting problematic content\pr{, such as hate speech,}  is a multifaceted and ever-changing task, influenced by social dynamics, user populations, diversity of sources, and evolving language. There has been significant efforts, both in academia and in industry, to develop annotated resources that capture various aspects of problematic content. Due to researchers' diverse objectives, the annotations are inconsistent and hence, reports of progress on detection of problematic content are fragmented. This pattern is expected to persist unless we consolidate resources considering the dynamic nature of the problem. We propose integrating the available resources, and leveraging their dynamic nature to break this pattern. In this paper, we introduce a continual learning benchmark and framework for problematic content detection comprising over 84 related tasks encompassing 15 annotation schemas from 8 sources. Our benchmark creates a novel measure of progress: prioritizing the adaptability of classifiers to evolving tasks over excelling in specific tasks. To ensure the continuous relevance of our framework, we designed it so that new tasks can easily be integrated into the benchmark. Our baseline results demonstrate the potential of continual learning in capturing the evolving content and adapting to novel manifestations of problematic content. \pr{\textit{\textbf{Warning}: this paper contains data that some readers may find offensive.}}

\end{abstract}

\section{Introduction}

Our social contexts continuously evolve and adapt to new situations, a trait that has enabled us to navigate through various challenges such as wars or pandemics. Peoples' expressions of hate, toxicity, and incivility, among other types of biases and prejudices, undergo adaptations in response to such changing circumstances. For instance, when there is a shift in the social or economic context, novel forms of hate speech emerge \citep{tahmasbi2021}. In such scenarios, fear and uncertainty contribute to the proliferation of stereotypical beliefs and the attribution of blame to particular groups \citep{velasquez2020hate, Cinelli_2020}. Even in stable social situations, differences in countries, contexts, and perspectives shape the boundaries of what is considered problematic content \pr{\citep{klonick2017new}}. 

The field of problematic content detection has produced an abundance of resources aiming to capture various aspects of this ever-changing phenomenon \citep{poletto2021resources, vidgen2020directions}. While the accumulation of such resources may appear to bring us closer to effectively addressing this problem, the static viewpoint adopted by each resource has resulted in heterogeneity among them, posing a significant challenge for integration of their knowledge into models. This heterogeneity has also caused fragmentation in progress reports on the automatic detection of problematic content. Therefore, it is crucial to establish a benchmark that integrates these annotated resources while capturing the dynamic nature of this problem. Such a benchmark would provide a more practical setting to test our models under stress and offer a new way to measure progress.
In this paper,  we introduce a continual learning benchmark and framework for problematic content detection comprising 84 related tasks encompassing 15 annotation schemas from 8 sources. By doing so, we present a novel perspective to address the problem of problematic content detection. Instead of focusing solely on specific aspects, such as the toxicity or incivility of a snapshot of a platform, we advocate for a dynamic formulation that builds on the ever-changing nature of problematic content. 
\begin{figure}
    \centering
    \includegraphics[width=\textwidth]{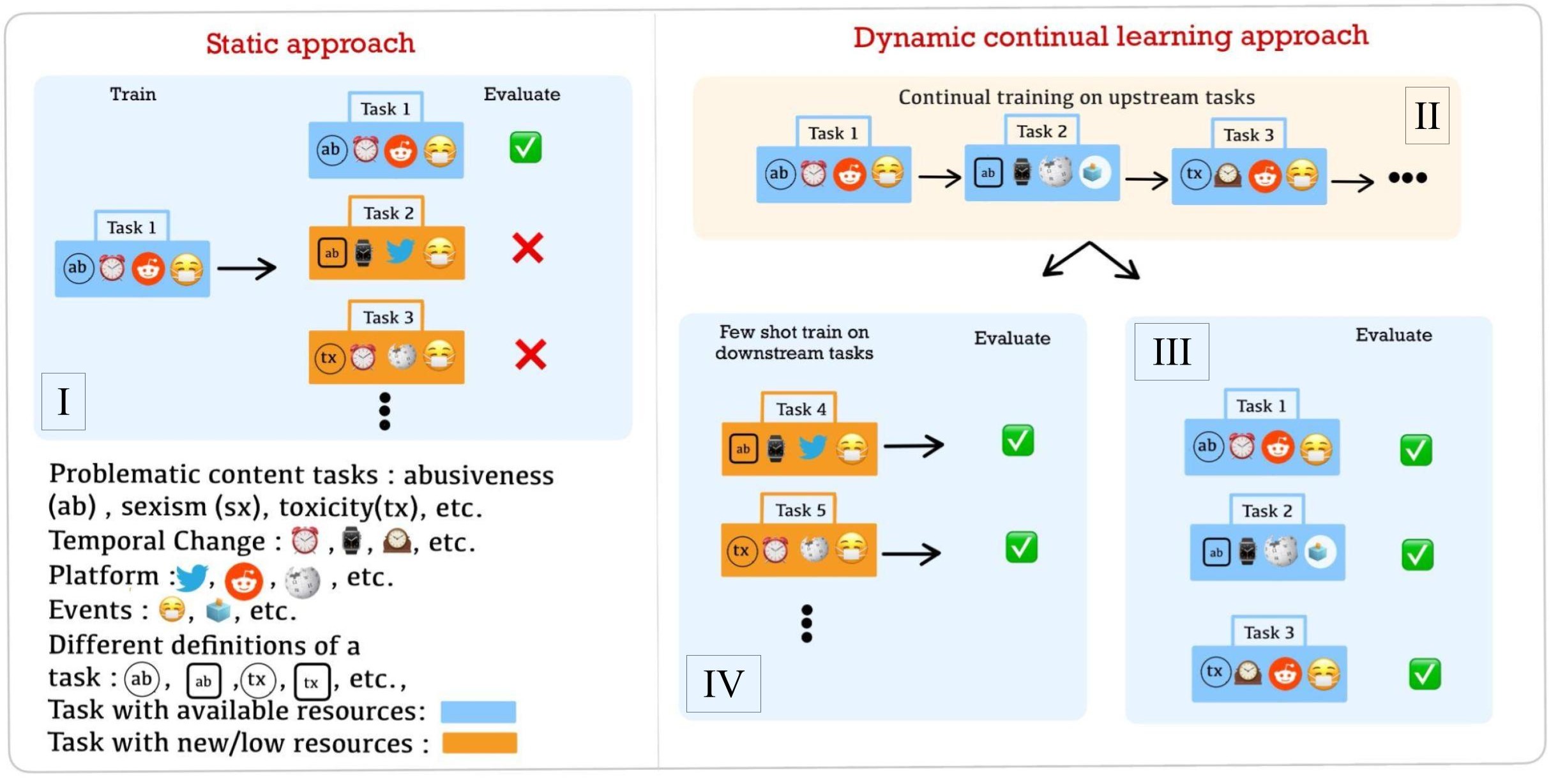}
    \caption{{Current static approaches (I) train and evaluate models on a fixed set of datasets. Our benchmark embraces the dynamic aspects of problematic content detection in two stages. The upstream trainig (II) and evaluation (III) where data is assumed to be coming in a stream, and downstream fewshot evaluation (IV) that measure models' generalization to novel forms of problematic content.  }}
    \label{fig:outline}
\end{figure}

Further, we propose a framework for identifying problematic content in a dynamic setting which satisfies the following two objectives: First, an optimal model should have the capability to acquire and retain knowledge about various types of problematic content. This capability is particularly crucial for effectively utilizing the diverse datasets that exist for detecting problematic content. We model this capability through a continual learning formulation, drawing inspiration from previous research \cite{robins1995catastrophic, de2019episodic, sun2019lamol}. Our models are designed to learn and understand the intricacies of problematic content by performing a diverse set of related tasks. Second, an optimal model should also have the ability to quickly learn and recognize new instances of problematic content, regardless of whether they appear on new platforms, in different languages, or target new groups. To assess and reward models that can adapt rapidly to emerging problematic content, we employ a few-shot evaluation benchmark on a separate set of related tasks, as suggested by recent work \cite{jin-etal-2021-learn-continually}.

Through these objectives, we establish criteria for an ideal model that can effectively handle the dynamic nature of problematic content. We define metrics and evaluations that capture these criteria, and we create a benchmark that accurately reflects the complexities of the problem. In constructing this benchmark, we integrate existing resources in the field, leveraging their strengths to develop a comprehensive framework for studying problematic content detection.

To validate the effectiveness of our proposed approach, we conduct extensive experimentation using diverse set of models and algorithms. Our evaluation focuses on detecting various types of problematic content such as hate speech, toxicity, and incivility, among others. Through these evaluations, we aim to showcase the strengths and limitations of each approach. The insights gained from our findings contribute to ongoing efforts in mitigating the harmful impact of problematic content on online platforms. This is discussed further in section \ref{sec:results} of our work. In sum, by addressing the dynamic nature of problematic content and embracing its complexities, our benchmark and experiments offer valuable insights, resources, and practical solutions for combating problematic content \footnote{Our benchmark and experiments are available at \url{https://github.com/USC-CSSL/Adaptable-Problematic-Content-Detection}}.

\section{Background}

\subsection{Problematic Content Detection}
Social media platforms offer individuals means to freely express themselves. However, certain features of social media, such as partial anonymity, which may promote freedom of expression, can also result in dissemination of problematic content. Researchers and social media companies recognize this issue and have developed various strategies to tackle it, including the use of automated systems to identify problematic content. Consequently, multiple definitions of problematic content have been proposed \citep{poletto2021resources}, encompassing specific areas like misogyny detection \citep[e.g., ][]{fersini2018overview}, to hate speech  \citep[e.g., ][]{kennedy2022introducing} and broader categories such as offensive language detection \citep[e.g., ][]{davidson2017automated}. Ideally, in order to foster more healthy and constructive online environments, such content detection systems should possess the capability to identify undesirable content irrespective of factors such as timing, specific linguistic form, or the social media platform used. However, recent studies have revealed limited generalizability of such systems, particularly in the context of hate speech detection \citep{yin2021towards, ramponi2022features}. \citet{yin2021towards} recognized that the scarcity of hate speech in sources poses a challenge to constructing datasets and models. They also acknowledged the difficulty in modeling implicit notions of problematic content. Integrating multiple datasets could potentially address both of these issues. By combining different datasets, the scarcity of problematic content would be reduced. Consequently, a model exposed to a greater variety of implicit notions would have a better ability to identify them.

\subsection{Multitask Learning for Problematic Content}

In recent years, multitask learning \citep{caruana1997multitask} has gained considerable attention as a promising approach for problematic content detection \cite{Kapil2021LeveragingMH, plaza2021multi, farha2020multitask, kapil2020deep, talat2018correction}. Multitask learning leverages the inherent relationships and shared characteristics among related tasks (e.g., hate speech, racism, sexism detection etc. in the context problematic content) to improve performance over a model that learns the tasks individually. By jointly training on multiple related tasks, the models can benefit from knowledge transfer and information sharing across different domains. This approach has shown potential in capturing the underlying nuances and contextual cues that are crucial for effectively detecting and addressing problematic content. \citet{kapil2020deep} conducted a comprehensive analysis on the potential benefits of simultaneously learning multiple problematic content tasks. They reported significant improvements including a 14\% and 12\% enhancement in macro $F_1$ over the state-of-the-art for offensive language detection \citep{zampieri2019semeval} racism and sexism detection \cite{waseem2016hateful} respectively.  Furthermore, empirical evidence shows the advantage of multitask learning in enhancing generalization and robustness. This advantage could potentially be due to the model's ability to learn common patterns and effectively differentiate between various forms of harmful language across different tasks \citep{mao2020multitask, zhou-etal-2019-improving, kapil2020deep}.  

Although multitask learning has demonstrated potential in the field of problematic content detection, it is not exempt from limitations. A significant drawback is the expense involved in retraining the model whenever a new task is introduced to the existing set. As the number of tasks grows, so does the complexity and computational resources needed for retraining. This becomes particularly challenging in the context of a dynamic landscape of problematic content, where new types of hate speech or toxic behavior emerge constantly. Multitask learning encounters various other challenges apart from computational complexity. These challenges include task interference, a phenomenon wherein the acquisition of multiple tasks concurrently can exert a detrimental impact on each other's learning processes, and catastrophic forgetting, which entails the loss of previously acquired knowledge when learning new tasks \citep{robins1995catastrophic, kirkpatrick2017overcoming, wu2023multi}.

\subsection{Continual Learning and Few Shot Generalization}

Continual learning is an approach that has emerged to address challenges like task interference, computational complexity, and catastrophic forgetting faced by multitask learning; instead of simultaneously learning all the tasks, continual learning models learn new tasks over time while maintaining knowledge of previous tasks \citep{robins1995catastrophic}. This incremental approach allows for efficient adaptation to new tasks while preserving the knowledge acquired from the previous tasks \citep{parisi2019continual}. By leveraging techniques such as parameter isolation, rehearsal, or regularization, continual learning mitigates catastrophic forgetting and ensures that the model retains its proficiency in previously learned tasks \citep{kirkpatrick2017overcoming,de2019episodic,wang2020efficient, schwarz2018progress}. Moreover, the capability to incrementally update the model alleviates the computational burden associated with retraining the entire multitask model every time new tasks are added. As a result, continual learning presents a promising approach to tackle the scalability and adaptability issues inherent in multitask learning. This framework becomes particularly attractive for tasks like hate speech detection, toxicity detection, incivility detection, and similar endeavors within a rapidly changing environment of problematic content. The only work in this space is \citet{qian-etal-2021-lifelong} which applies continual learning to detect hate speech on Twitter. However, 
their focus is limited to a single definition of hate speech and they analyze a single snapshot of Twitter data. Consequently, their approach does not fully account for the dynamic nature of problematic content across the internet.

\section{Continual Learning Benchmark for Problematic Content Detection}
\label{sec:benchmark}

\subsection{Problem Formulation}
\label{sec:problem_formulation}

Our objective in creating this benchmark is to develop models that are not only agile in detecting new manifestations of problematic content but are also capable of accumulating knowledge from diverse instances across different time periods and platforms. Such models should possess the ability to rapidly learn and identify new manifestations of problematic content on novel platforms, even when only limited data is available. As time progresses, we anticipate a natural increase in the availability of resources for problematic content detection. Therefore, to encourage building models that leverage this increase in resources, we consider the existing resources as a continuous stream of incoming data. In this context, we make the assumption that there exists a problematic content detection model denoted as $f$, which undergoes continual learning on a stream of problematic content detection tasks ($T^u = [T_1^u, \ldots, T_{N_u}^u]$) over time. We refer to this set of tasks as \textit{upstream} tasks. In addition to accumulating knowledge from the stream of tasks, this continual learning model should be able to rapidly generalize its knowledge to numerous related unseen tasks \citep{jin-etal-2021-learn-continually}. We formulate this ability as a few-shot learning problem over a separate set of tasks $T^d = [T_1^d, \ldots, T_{N_d}^d]$, referred to as \textit{downstream} tasks.

\subsection{Training and Evaluation}
\label{sec:train-eval}
During the continual learning stage, the model encounters a sequentially ordered list of $N_u$ upstream tasks: $[T_1^u, \ldots, T_{N_u}^u]$, where each task has its own distinct training and test sets. To evaluate the few-shot learning capability of the sequentially trained model $f$, we proceed to adapt it to a collection of $N_d$ few-shot tasks individually represented as ${T_i^d}$. In this scenario, each unseen task is associated with only a small number of training examples.

\pr{For evaluation purposes, a task is considered ``new'' if the model hasn’t been exposed to labels from that task. This applies to the $i_{th}$ upstream task ($T_i^u$) in the upstream training process before the model’s upstream training reaches $T_i^u$, as well as to all downstream tasks (Figure \ref{fig:outline})} The paucity of problematic content online results in most datasets used in this work to be quite unbalanced (see supplementary materials for details). In unbalanced datasets, AUC is often preferred over $F_1$ score \citep{bradley1997use}. Hence, we chose AUC as our primary evaluation metric for both the upstream training and downstream adaptation processes. \pr{To enable fair comparisons, we used a fixed set of held-out test data for all models.} Below we outline the specific measures we employ to characterize the desired attributes of each model.

\noindent\textbf{Few-Shot Performance}
To assess the model's few-shot generalization ability, we evaluate the continually trained model $f$ on unseen tasks by individually fine-tuning it for each task $T_i^v$ using a few annotated examples. The few-shot AUC for task $T_i^d$ is denoted as $AUC_{i}^{F S}$, 
and we report the average few-shot AUC ($AUC^{FS}$) across all downstream tasks.

\noindent\textbf{Final Performance}
To assess the accumulation of knowledge in upstream tasks, we  evaluate the AUC of $f$ at the end of the continual learning over upstream tasks. This evaluation allows us to determine the extent to which model $f$ forgets the knowledge pertaining to a specific task once it acquires the ability to solve additional tasks. We report the average AUC over all upstream tasks.

\noindent\textbf{Instant Performance}
To assess the effect of learning a sequence of upstream tasks on learning a new upstream task, we evaluate the AUC of $f$ on task $T_i^u$ right after the model is trained on $T_i^u$. We report the average of instant performance across all upstream tasks.

\subsection{Datasets} \label{datasets}

We have selected datasets for our benchmark based on the following criteria: (1)  must be related to problematic content detection, (2) must be in English, and (3)  must include a classification task (or a task transformable into classification). We aimed to use datasets that span different sources and time periods, and rely on different definitions of problematic content. Even though we currently focus on one language, the dynamic nature of our formulation easily allows for expansion of this benchmark to other languages \pr{(see \S \ref{limitation} for more details)}.  Our benchmark currently covers data from 8 different sources, namely, Twitter, Reddit, Wikipedia, Gab, Stromfront, chat dialogues, and synthetically generated text.
These datasets cover a wide range of definitions of problematic content, from focused definitions such as sexism and misogyny to broader definitions such as toxicity.

For all datasets, we use the original train/test/dev splits when available, otherwise split the data 80/10/10 randomly. We briefly discuss each dataset below; [U] denotes upstream datasets and [D] is used for datasets used in downstream.

\noindent\textbf{Call Me Sexists, But} \citep[CMSB;][]{liakhovets2022transfer} [D]
Consists of 6,325 tweets from two sources: 1) Twitter data that was previously annotated for sexism and racism \citep{waseem2016hateful}, and 2) Twitter data collected between 2008 and 2019 using the phrase ``call me sexist, but.'' Each tweet in the dataset is labeled for sexist content and sexist phrasing, with both being single-choice options, using a set of guidelines derived from psychological scales.

\noindent\textbf{US-election} \citep{grimminger-klinger-2021-hate} [D]
Consists of 3000 tweets, covering hate speech and offensive language, which were collected during the six weeks prior to the 2020 presidential election, until one week after the election. Each tweet was annotated for being hateful/non-hateful without considering whether the target is a group or a single person.

\noindent\textbf{Misogyny Detection} \citep[misogyny;][]{guest2021expert} [D]
Contains 6567 Reddit Posts from 34 subreddits identified as misogynistic from Feb to May 2020 annotated with a three level hierarchical taxonomy. We only use the top level annotations which are binary labels for misogynistic content. 


\noindent\textbf{Contextual Abuse Dataset} \citep[CAD;][]{vidgen2021introducing} [U]
Cosists of 25k Reddit posts collected from 16 Subreddits more likely to contain a diverse range of abusive language, and focused on taking the context of the conversations into account. A hierarchical annotation schema is proposed which takes the context of the conversation into account; Level 1: abusive, non-abusive, and Level 2: for abusive (i) identity-directed, (ii) affiliation-directed and (iii) person-directed. In our benchmark, we use the three labels from the second level to stress test models' ability in learning variations of abuse.


\noindent\textbf{Ex-Machina: Personal Attacks at Scale} \citep[Personal attack;][]{wulczyn2017ex} [U]
Includes 100k annotated comments from a public dump of Wikipedia from 2004-2015 and annotators were asked to label comments that contain personal attack or harassment in addition to some finer labels about the category of attack or harassment. We included the detecting personal attacks, quoted personal attacks (QA), and personal attack targeted at third party (TPA) as separate tasks in our benchmark.

\noindent\textbf{Unhealthy Comment Corpus} \citep[UCC;][]{price-etal-2020-six} [U]
Consists of 44,355 comments collected from the Globe and Mail news site. Every comment is annotated according to a two-level hierarchy; Level 1: healthy or unhealthy. Level 2: binary labels indicating the presence or absence of six specific unhealthy subattributes: (i) hostility, (ii) antagonism, (iii) insults, (iv) provocation, (v) trolling, (vi) dismissiveness, (vii) condescension, (viii) sarcasm, and (ix) generalization.

\noindent\textbf{The Gab Hate Corpus} \citep[GHC;][]{kennedy2022introducing}[U]
Contains 27,665 posts from \url{Gab.com}, spanning January, 2018 to October, 2018, annotated based on a typology for hate speech derived from definitions across legal precedent. Posts were annotated for Call for Violence (CV), Human degradation (HD), Vulgarity and/or Offensive language (VO) and whether a comment contains explicit or implicit language.

\noindent\textbf{Stormfront} \citep{de2018hate} [D]
Includes a 10,568 sentences collected from 22 sub-forums of \url{Stormfront.org} spanning from 2002 to 2017. Each sentence has been classified as containing hate or not depending on whether they meet the following three premises: ``a) deliberate attack, b) directed towards a specific group of people, and c) motivated by aspects of the group’s identity.''

\noindent\textbf{Dialogue Safety} \citep{miller2017parlai,xu2021bot} [D]
The Dialogue Safety dataset includes five datasets in the domain of dialogue safety. Three datasets, namely ParlAI single standard, ParlAI single adversarial, and ParlAI multi, are sourced from ParlAI \citep{miller2017parlai}. The other two datasets, BAD2 and BAD4, are from Bot-Adversarial Dialogue \citep{xu2021bot}. The ParlAI datasets consist of 30,000 samples, while the BAD datasets consist of 5,784 samples. Conversations in the BAD dataset can span up to 14 turns, and following \cite{xu2021bot}, we consider the last two and four utterances of the conversation (BAD2 and BAD4)  in our benchmark. All dialogue safety datasets provide toxic or safe labels.

\noindent\textbf{Dygen} \citep{vidgen-etal-2021-learning} [hate U, rest D] 
Consists of 41,255 samples dynamically generated using the human-and-model-in-the-loop setting to train more robust hate detection models. The authors collected four rounds data using \emph{Dynabench} \citep{kiela2021dynabench}, and annotated each sample hierarchically; Level 1:  binary hate/non-hate label, Level 2:  subclasses of hate (i.e., derogation, animosity, threatening language, support for hateful entities and dehumanization) and 29 target identities (e.g., immigrant, muslim, woman, etc.). We use Level 1 for upstream training and Level 2 for downstream adaptation.

\noindent\textbf{Hatecheck} \citep{rottger-etal-2021-hatecheck} [D] 
Contains of 3,728 synthetically generated sentences motivated by 29 hate speech detection model functionalities; 18 of these functionalities test for hateful content and cover distinct expressions of hate, and the other 11 functionalities test for non-hateful content and cover contrastive non-hate. 

\noindent\textbf{Multitarget-CONAN} \citep[CONAN;][]{fanton2021human} [D] 
Consists of 5003 samples of hate speech and counter-narrative pairs targeting different target groups (LGBTQ+, Migrants, Muslims, etc.) created using human-in-the-loop methodology, in which the generative language model generates new samples and, after confirmation by expert annotators, would get added to the dataset. In our benchmark we included detection of hate speech toward each target group as a separate task.


\noindent\textbf{Civil-comments} \citep{dixon2018measuring} [U]
Includes two million comments from the Civil Comments platform which is annotated by human raters for various toxic conversational attributes. Each comment has a toxicity label and several additional toxicity subtype attributes which are severe toxicity, obscene, threat, insult, identity attack, sexual explicit. 


 \noindent\textbf{Twitter Abusive} \citep[Abusive;][]{founta2018large} [U]
Contains 80k tweets from March to April 2017 annotated for multiple fine-grained aspects of abuse, namely, offensiveness, abusiveness, hateful speech, aggression, cyberbullying, and spam.

\noindent\textbf{Large-Scale Hate Speech Detection with Cross-Domain Transfer} \citep[hate;][]{toraman-etal-2022-large} [U]
Includes 100k tweets from 2020 and 2021, each annotated by five annotators for hate speech. Tweets are labeled as hate if ``they target, incite violence against, threaten, or call for physical damage for an individual or a group of people because of some identifying trait or characteristic.''


\section{Models and Methods}

\subsection{Models}
\label{sec:models}

 We represent all tasks in a consistent binary classification format and conduct our experiments using a pretrained language model, specifically \textbf{BART-Base} \citep{lewis2020bart}. In addition to fine-tuning all the model weights of BART-Base, we also explore two other variations:
 
 \noindent\textbf{BART-Adapter:} We experiment with Adapter training  \textbf{\citet{houlsby2019parameter}}. In addition to the classification head, adapter training only trains parameters of Adapters, which are two-layer MultiLayer Perceptrons (MLPs) inserted after each layer of BART. 
 
 \noindent\textbf{BART-HNet:} Following \citep{jin-etal-2021-learn-continually}, we explore using hypernetworks (BART-HNet). The hypernetwork ($h$) accepts a task representation $z$ as input and generates model parameters for a separate prediction model, denoted as $f$, in order to address the specific task at hand.

\subsection{Upstream Training}
\noindent\textbf{Single Task Learning}
 We \pr{finetune} a pretrained model on each of the upstream tasks $T_i^u$ separately. Note that this model completely ignores the sequential order imposed to our upstream tasks and will, therefore, serve as a baseline for evaluating the performance of our base model on all upstream tasks without any knowledge transfer. 

\noindent\textbf{Sequential Finetuning (\textbf{Vanilla})}
We also finetune a pretrained model on the sequence of upstream tasks $[T_1^u, \ldots, T_{Nu}^u]$ without any continual learning algorithms. Previous research suggests that this model will suffer from catastrophic forgetting \citep{robins1995catastrophic}. Comparing the final performance of this model with a continual learning algorithm will give us a measure of the ability of these algorithms in knowledge accumulation. 

\noindent\textbf{Multitask Learning (\textbf{MTL})}
To assess the upper bound of knowledge accumulation on the set of upstream tasks 
we \pr{finetune} a pretrained model with multitask learning on all upstream tasks. Multitask learning is implemented via hard parameter sharing (see supplementary materials for implementation details).

\noindent\textbf{Continual Learning}
Finally, we \pr{finetune} a model continually on a sequence of upstream tasks $[T_1^u, \ldots, T_{Nu}^u]$. This model should ideally be able to 1. use knowledge from previous tasks to learn a new upstream task, and 2. retain knowledge of the seen upstream tasks.

We experiment with two continual leanrning algorithms:

\noindent\textbf{Bi-level Hypernetworks for Adapters with Regularization} \citep[\textbf{BiHNet-Reg}:][]{jin-etal-2021-learn-continually}. This model is specifically designed to enhance the generation of adapter weights by optimizing bi-level task representations. Its primary objective is to address two important challenges: mitigating catastrophic forgetting and enhancing the overall generalizability of the model. Towards the first challenge,  regularization is imposed on the generated adapters. To improve generalization this model learns two representations for each task task; one for high-resource settings and one for few-shot cases.

\noindent\textbf{Elastic Weight Consolidation} \citep[\textbf{EWC}:][]{kirkpatrick2017overcoming}: leverages the principles of Bayesian inference, suggesting a method that selectively slows down learning on the weights important for previous tasks. The model retains old knowledge by assigning a larger penalty to changes in crucial parameters, effectively making them ``elastic''. 

\subsection{Downstream Adaptation}

An ideal model for problematic content detection should be able to learn its new manifestations quickly. Therefore, we evaluate our models' ability on learning unseen datasets of problematic content using only a few examples. We report the performances using $k=16$ shots (further analysis into the effect of number of shots on model performances is provided in supplementary materials).

\section{Experiments}
\label{sec:experiments}

\begin{figure}
    \centering
    \includegraphics[width=\textwidth]{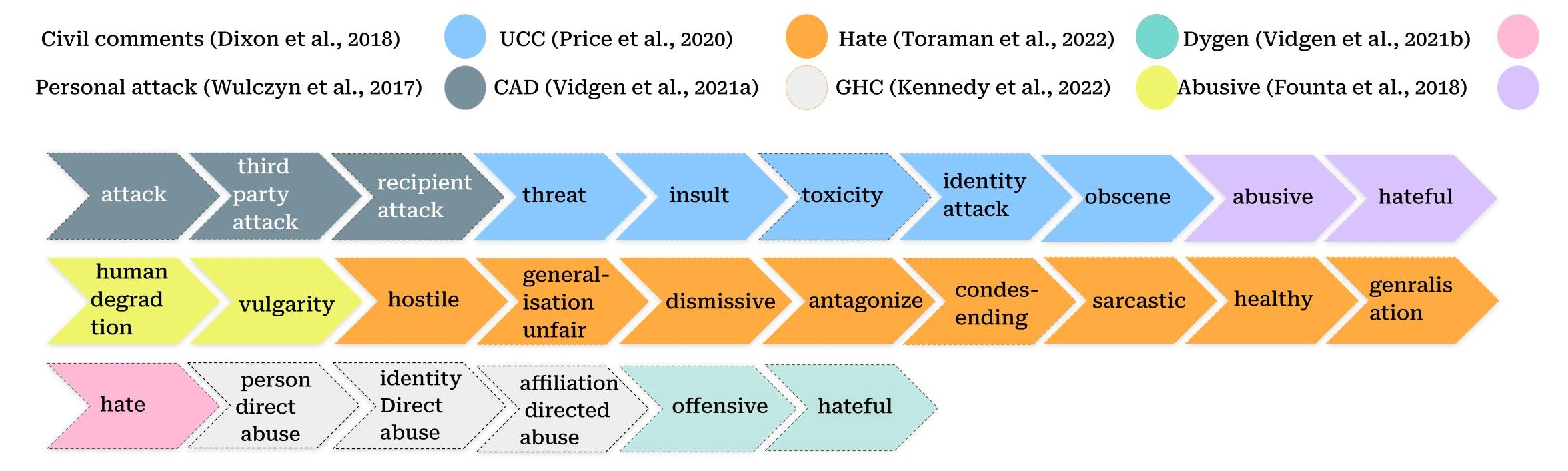}
    \caption{{Sequence of upstream tasks in the experiment with chronological task order. Note that datasets are ordered according to the earliest publication date of the data and tasks (i.e., labels) within each dataset are ordered randomly.}}
    \label{fig:chrono-experiment}
\end{figure}


Most of the datasets in our benchmark include annotations for various aspects of problematic content (e.g., the UCC dataset includes labels for antagonism, insults, etc.). To ensure flexibility, we treated each label as a separate task. Our rationale behind this approach is rooted in the likely possibility that we will need to introduce additional labels to the existing set in the future. To accommodate potential future updates to the label taxonomy, it is preferable to have models that can quickly adapt and learn new labels. We include each label as a task if the training split includes at least 100 positive samples.
In order to minimize the exchange of information between the upstream and downstream tasks, across all our datasets with the exception of Dygen, we categorized all tasks within the dataset as either upstream or downstream. Our selection of larger datasets for the upstream tasks was driven by both the data requirements of upstream training and the fact that larger datasets typically encompass a broader range of problematic content. This decision enables the model to accumulate knowledge on general notions of problematic content, which aligns with our objectives. \pr{Subsequently, we assigned tasks as downstream that 1. had limited labeled data, and 2. had minimal overlap (e.g., same domain or labels) with the upstream tasks.} \pr{
To assess the efficacy of our proposed framework in practical scenarios, we ran an additional experiment where we ordered the upstream tasks \textit{chronologically}. Specifically, we used the earliest publication date of each dataset as the temporal reference point to order the upstream datasets. Note that each dataset consists of multiple labels (i.e., tasks). Since we don’t have any information about the temporal order of tasks within datasets, we chose this order at random. This experiment allowed us to capture the evolution of the research landscape on problematic content detection, thereby providing a more nuanced understanding of the progress of model performance over time. Figure \ref{fig:chrono-experiment} shows the order of upstream tasks in this experiment. }

\begin{table}[H]
    \centering
    
\begin{tabular}{lcccr}
\toprule
model & Fewshot-F1 & Fewshot-AUC &  $\Delta$ Fewshot-AUC \\
\midrule
\pr{BART-Adapter-Vanilla} & \pr{0.256353} & \pr{0.764620} &  \pr{-}  \\
\pr{BART-BiHNet-Vanilla} & \pr{0.270319} & \pr{0.771721} & \pr{-}  \\
\pr{BART-BiHNet-Reg} & \pr{0.298919} & \pr{0.818512} & \pr{+0.046791}  \\
\pr{BART-BiHNet-EWC} & \pr{0.257523} & \pr{0.765808} &  \pr{-0.005913}  \\
\pr{BART-Adapter-Multitask} & \pr{0.288102} & \pr{0.816277} & \pr{+0.051657}  \\
\pr{BART-BiHNet-Multitask} & \pr{0.257100} & \pr{0.795745} &  \pr{+0.024024}  \\
\bottomrule
\end{tabular}
\caption{ {Fewshot performance for the models on the chronological experiment} }
    \label{tab:chrono-fs}
\end{table}
\begin{table}[H]
    \centering
    
\begin{tabular}{lcc}
\toprule
model & Final-F1 & Final-AUC  \\
\midrule
\pr{BART-Adapter-Vanilla} & \pr{0.130313} & \pr{0.517685}  \\
\pr{BART-BiHNet-Vanilla} & \pr{0.098584} & \pr{0.617468}  \\
\pr{BART-BiHNet-Reg} & \pr{0.271746} & \pr{0.791544}  \\
\pr{BART-BiHNet-EWC} & \pr{0.074141} & \pr{0.676287}  \\
\pr{BART-Adapter-Multitask} & \pr{0.382802} & \pr{0.872739}  \\
\pr{BART-BiHNet-Multitask} & \pr{0.318443} & \pr{0.833905}  \\
\bottomrule
\end{tabular}
\caption{{Final performance for the models on the chronological experiment} }
    \label{tab:chrono-final}
\end{table}

\begin{table}[H]
    \centering
    
\begin{tabular}{lcccr}
\toprule
model & Instant-F1 & Instant-Auc & $\Delta$ Instant-AUC \\
\midrule
\pr{BART-Adapter-Vanilla} & \pr{0.402099} & \pr{0.882196} & \pr{-}  \\
\pr{BART-BiHNet-Vanilla} & \pr{0.412636} & \pr{0.878479} & \pr{-}  \\
\pr{BART-BiHNet-Reg} & \pr{0.399602} & \pr{0.881692} & \pr{+0.003213}  \\
\pr{BART-BiHNet-EWC} & \pr{0.403327} & \pr{0.880995} & \pr{+0.002516}  \\
\bottomrule
\end{tabular}
\caption{{Instant performance for the models on the chronological experiment }}
    \label{tab:chrono-instant}
\end{table}

\pr{To show the efficacy of our proposed continual learning approach in adapting to any scenario, we have also experimented with randomly ordering all the upstream tasks.} More details on this experiment can be found in the supplementary materials.

\section{Results}
\label{sec:results}

\noindent\textbf{Single Task Baseline:}
To determine the learning capabilities of each model architecture for different tasks \ref{sec:models}, we \pr{finetune} a classifier from each architecture on each task. The average fewshot, final, and instant performance of BART-Adapter-Vanilla, and BART-HNet-Vanilla is presented in the first two rows of tables \ref{tab:chrono-fs}, \ref{tab:chrono-final}, \ref{tab:chrono-instant} respectively. We see the largest gap in performance for these models on the final performance metrics. This can be attributed to BiHNet's meta learning capabilities. The details of model performances on each task can be found in supplementary material.

\noindent\textbf{Multitask Upperbound:}
In scenarios where there are no adversarial tasks, multitask learning is often used as an empirical upper bound for continual learning algorithms. The last two rows of tables \ref{tab:chrono-fs} and \ref{tab:chrono-final} and  the few shot and final evaluation of multitask models. Note that since these models see all tasks at the same time, instant performance is not defined for them.

\noindent\textbf{Does the collection of problematic content tasks help with learning new upstream tasks?}
In other words, do the models benefit from upstream training when learning a new task with substantial amount of annotated data available? To answer this question, compare the instant performance of a CL model on $T_i^u$ with a pretrained model finetuned on just  $T_i^u$. Our results ($\Delta$ Instant AUC) show evidence of slight positive transfer, however, the magnitude of this transfer is negligible.

\noindent\textbf{Does continual learning improve knowledge retention?}
The final AUC values, as shown in  Table \ref{tab:chrono-final}, indicate the models' ability to retain knowledge from a sequence of tasks at the end of training. Our results suggest that all continual learning variations outperform naive training. Most notably, BiHNet-Reg outperforms BiHNet-Vanilla by almost 18\% in AUC, indicating its potential to mitigate catastrophic forgetting, while falling only 4\% short of the multitask counterpart.

\noindent\textbf{Does upstream learning help with generalization to new manifestations of problematic content?}
Comparing the single-task baselines with continual and multitask learning, our results (Table \ref{tab:chrono-fs}) demonstrate a noteworthy improvement in models' generalization ability as a result of upstream training. \pr{Specifically, BiHNet-Reg shows remarkable generalization ability in fewshot settings, outperforming the BiHNet-Vanilla by nearly 5\% in AUC.}

\section{Discussion and Conclusion}
In conclusion, we propose a continual learning benchmark and approach for detecting problematic content, that realizes its dynamic and adaptable nature. We define essential characteristics of an ideal model and create a continual learning benchmark evaluation metrics to capture the variability in problematic content. Our benchmark has two key components: First, an upstream sequence of problematic tasks over which we measure a model's ability in accumulating knowledge, and second, a separate set of downstream few-shot tasks on which we gauge a model's agility in learning new manifestations of problematic content. Our experiments clearly demonstrate the effectiveness of this formulation, particularly in its ability to adapt to new types of problematic content. To ensure the benchmark remains dynamic and up-to-date, we have designed it with continuous updates in mind; our benchmark's flexible implementation allows for seamless repositioning of tasks as either upstream or downstream. We encourage the community to actively contribute to and expand this benchmark, as it serves as a collaborative platform for advancements in the field.

\section{Limitation and Negative Societal Impact}
\label{limitation}
\pr{The social science examination of the evolution of problematic content carries its own importance and follows a dedicated line of inquiry. Due to space constraints, we have not provided an exhaustive discussion of this subject. We recommend referring to \cite{klonick2017new, atlanticcouncil2023} for a comprehensive overview of this area.} The benchmark under discussion is currently designed only for English language content, neglecting the challenges posed by problematic content in other languages and cultures. \pr{Our design, however, allows for an easy expansion of the benchmark to include other languages. We have outlined the procedure to expand the benchmark on the accompanying repository and encourage the community to contribute to the benchmark. } Though it presents a new measure of progress and baseline results, further investigations and extensive experimentation are needed to fully evaluate the potential of continual learning in detecting evolving problematic content. The study's approach, predominantly using majority label datasets, potentially leads to bias and overgeneralization in detecting problematic content, given the inherent subjectivity of such content influenced by cultural norms, individual sensitivities, and societal changes over time. The effectiveness of this benchmark could significantly vary due to the diversity of sources and annotation schemas, potentially leading to cultural bias and an overreliance on AI for content detection, thereby neglecting the importance of nuanced human moderation. \pr{Future work can explore the potential considering this subjectivity under our continual learning framework.} Moreover, the benchmark opens possibilities for misuse, including training models to generate problematic content or designing adversarial attacks, where malicious actors can exploit the understanding of detection systems to craft content that evades detection.

\pr{Datasets used in this benchmark may have a high prevalence of problematic content targeting certain social groups. This, in turn, can lead subsequent models to produce unfair outcomes, such as higher false positive rates for the aforementioned groups \cite{dixon2018measuring, wiegand2019detection}. Recently, various methods have been proposed to mitigate these biases, such as those by \citet{mostafazadeh-davani-etal-2021-improving, kennedy2020contextualizing}. Future research could examine the extent of biases' influence on the model within our framework and the effectiveness of the mentioned techniques in mitigating them. Moreover, Some datasets may hold personally identifiable information or data from which individual details can be inferred. To address this concern, we suggest applying Google's DLP, a tool designed to scan and classify sensitive data, to the datasets. Another concern in research on problematic content detection is potential misuse for censorship. However, we emphasize that, in contrast to private methods concealed behind corporate doors, an open-access or academic approach to detecting problematic content fosters transparency. This allows the public to understand and critique the detection criteria. Such transparency ensures accountability, given that academic methods frequently undergo peer review and public scrutiny, thereby addressing biases and mistakes.}

\begin{ack}
This research was supported by NSF CAREER BCS-1846531 and DARPA INCAS HR001121C0165. The views and conclusions contained herein are those of the authors and should not be interpreted as necessarily representing the official policies, either expressed or implied, of DARPA, or the U.S. Government. The U.S. Government is authorized to reproduce and distribute reprints for governmental purposes notwithstanding any copyright annotation therein.
\end{ack}

\bibliography{bibliography}

\section*{Checklist}


\begin{enumerate}

\item For all authors...
\begin{enumerate}
  \item Do the main claims made in the abstract and introduction accurately reflect the paper's contributions and scope?
    \answerYes{}
  \item Did you describe the limitations of your work?
    \answerYes{Section \ref{limitation}}
  \item Did you discuss any potential negative societal impacts of your work?
    \answerYes{Section \ref{limitation}}
  \item Have you read the ethics review guidelines and ensured that your paper conforms to them?
    \answerYes{}
\end{enumerate}

\item If you are including theoretical results...
\begin{enumerate}
  \item Did you state the full set of assumptions of all theoretical results?
\answerNA{}	
\item Did you include complete proofs of all theoretical results?
\answerNA{}
\end{enumerate}

\item If you ran experiments (e.g. for benchmarks)...
\begin{enumerate}
  \item Did you include the code, data, and instructions needed to reproduce the main experimental results (either in the supplemental material or as a URL)?
    \answerYes{footnote on page 3}
  \item Did you specify all the training details (e.g., data splits, hyperparameters, how they were chosen)?
    \answerYes{Included in supplementary material}
	\item Did you report error bars (e.g., with respect to the random seed after running experiments multiple times)?
    \answerYes{Included in supplementary material}
	\item Did you include the total amount of compute and the type of resources used (e.g., type of GPUs, internal cluster, or cloud provider)?
    \answerYes{Included in supplementary material}
\end{enumerate}

\item If you are using existing assets (e.g., code, data, models) or curating/releasing new assets...
\begin{enumerate}
  \item If your work uses existing assets, did you cite the creators?
    \answerYes{}
  \item Did you mention the license of the assets?
    \answerYes{Information provided along with how to obtain the data on the repository - link to repository provided on footnote of page 3}
  \item Did you include any new assets either in the supplemental material or as a URL?
    \answerYes{The benchmark and code are released as a part of the project repository - link provided on footnote of page 3}
  \item Did you discuss whether and how consent was obtained from people whose data you're using/curating?
    \answerYes{Information provided along with how to obtain the data on the repository - link provided on footnote of page 3}
  \item Did you discuss whether the data you are using/curating contains personally identifiable information or offensive content?
    \answerYes{All datasets are about problematic content and we discuss each dataset briefly in section \ref{sec:benchmark}}
\end{enumerate}

\item If you used crowdsourcing or conducted research with human subjects...
\begin{enumerate}
  \item Did you include the full text of instructions given to participants and screenshots, if applicable?
    \answerNA{}
  \item Did you describe any potential participant risks, with links to Institutional Review Board (IRB) approvals, if applicable?
    \answerNA{}
  \item Did you include the estimated hourly wage paid to participants and the total amount spent on participant compensation?
    \answerNA{}
\end{enumerate}

\end{enumerate}

\section{Supplementary Material}
\label{sec:supp}

\subsection{Hardware and Runtimes}
Experiments were conducted on Nvidia Quadro 6000 GPUs with Cuda version 11.4. Each upstream training for 26 tasks takes around 12 hours, and few-shot training and evaluation for all 58 downstream tasks for a single model takes around 6 hours to complete.

\subsection{Data Sources and License Information}

All of the datasets used in this benchmark are publicly available \pr{for research purposes. Table \ref{tab:license} provides license information for all datasets used in this benchmark.}
The instructions in our Github repository\footnote{\url{https://github.com/USC-CSSL/Adaptable-Problematic-Content-Detection}} offer a clear guide on how to create a local copy of all the datasets used in our benchmark.

\begin{longtable}{|c|c|p{4.5cm}|}
    \hline
    \textbf{Name} & \textbf{License} & \textbf{Source} \\
    \hline
    \endfirsthead
    
    \hline
    \textbf{Name} & \textbf{License} & \textbf{Source} \\
    \hline
    \endhead

    \pr{UCC and Ex Machina} & \pr{CC-BY-SA} & \href{https://en.wikipedia.org/wiki/Wikipedia:Copyrights#Contributors'_rights_and_obligations}{https://en.wikipedia.org/wiki/  Wikipedia} \\
    \hline
    \pr{Civil Comments Corpus} & \pr{CC0} & \href{https://www.kaggle.com/competitions/jigsaw-unintended-bias-in-toxicity-classification/data}{https://www.kaggle.com/  competitions/jigsaw-unintended-bias-in-toxicity-classification/data}\\
    \hline
    \pr{Misogyny Detection} & \pr{MIT} & \url{https://github.com/ellamguest/online-misogyny-eacl2021} \\
    \hline
    \pr{CAD} & \pr{CC-By Attribution  4.0 International} &  \url{https://zenodo.org/record/4881008} \\
    \hline
    \pr{DYGEN} & \pr{CC By 4.0} & footnote of the first page of the paper: \url{https://dl.acm.org/doi/pdf/10.1145/3580305.3599318} \\
    \hline
    \pr{HateCheck} & \pr{CC By 4.0} & \url{https://github.com/paul-rottger/hatecheck-data/blob/main/LICENSE} \\
    \hline
    \pr{CONAN} & \pr{``resources can be used}  & \\ & 
    \pr{for research purposes''} & \url{https://github.com/marcoguerini/CONAN} \\
    \hline
    \pr{Stormfront} & \pr{CC-BY-SA} & \url{https://github.com/Vicomtech/hate-speech-dataset} \\
    \hline
    \pr{GHC} & \pr{CC-By Attribution 4.0 International} & The GHC is available on the Open Science Framework (OSF, \url{https://osf.io/edua3/}), and license is discussed in detail in section 4 of the paper \\
    \hline
    \pr{CMSB} & \pr{CC BY-NC-SA 4.0} & \url{https://data.gesis.org/sharing/#!Detail/10.7802/2251} \\
    \hline
    \pr{Large-Scale Hate} & & \\\pr{Speech Detection} & &  \\ \pr{with Cross-Domain Transfer} & \pr{CC-BY-SA 4.0} & \url{https://github.com/avaapm/hatespeech/blob/master/LICENSE} \\
    \hline
    \pr{US Election} & \pr{data is publicly available} & \url{https://www.ims.uni-stuttgart.de/forschung/ressourcen/korpora/stance-hof/} \\
    \hline
    \pr{Dialogue Safety} & \pr{MIT} & \url{https://github.com/facebookresearch/ParlAI/blob/main/LICENSE} \\
     
    \hline
    \pr{Twitter Abusive} & \pr{CC-By Attribution 4.0 International} & \url{https://zenodo.org/record/2657374} \\
    \hline
    \caption{ {License information for all dataset used in benchmark. According to this information all datasets can be used for research purposes}}
    \label{tab:license}
\end{longtable}



\begin{table}[H]
    \centering
    \begin{tabular}{|l|p{11cm}|}
\hline
\textbf{Dataset} & \textbf{Labels}  \\
\hdashline
Abusive & abusive (2763); hateful (503); total (8597) \\
\hdashline
CAD & affiliation directed abuse (242) ; identity directed abuse (514); person directed abuse (237); total (5307) \\
\hdashline
Dygen & hate (2268); total (4120) \\
\hdashline
GHC & human degradation (491); vulgarity (369); total (5510) \\
\hdashline
Gate & hateful (170); offensive (1247); total (10207) \\
\hdashline
Civil comments & identity attack (687); insult (5776); obscene (543); threat (221); toxicity (7777); total (97320) \\
\hdashline
Personal attack & attack (3056) ; recipient attack (1999) ; third party attack (204); total (23178) \\
\hdashline
UCC & antagonize (203);  condescending (269) ; dismissive (150) ;  generalisation (96) ;  generalisation unfair (91) ; healthy (320) ; hostile (108) ; sarcastic (201) ; total (4425) \\
\hline
\end{tabular}
\caption{Number of label occurrences in upstream tasks test sets. }
\label{tab:upstream_details}
\end{table}

\begin{table}[H]
    \centering
    \begin{tabular}{|l|p{11cm}|}
\hline
\textbf{Dataset} & \textbf{Labels} \\
\hline
Dygen & Black men (7); African (8); Muslim women (10); Asylum seekers (13); Asians (15); Indigenous people (18); Gender minorities (21); Chinese people (25); Foreigners (26); Black women (27); Travellers (27); Non-whites (28); Mixed race (30); Gay women (31); East Asians (32); South Asians (32). Gay men (34); support (35); Arabs (45); threatening (48); Refguees (51); dehumanization (70); People with disabilities (79); Gay people (81); Immigrants (81); Trans people (90); Jewish people (111); Muslims (126); Black (211); animosity (315); derogation (1036); total (3009)\\
\hdashline
CONAN & disabled (22);  jews (59); muslims (134); migrant (96); woman (67); LGBT (62); people of color (35); total (501)\\
\hdashline
Hatecheck & trans (42); black (44); immigrants (45); muslims (47); gay (48); disabled (50); women (60); hate (117); total (373) \\
\hdashline
single & \multirow{3}{*}{toxic (300); total (3000)} \\
adversarial &  \\
multi &  \\
\hdashline
BAD2 & \multirow{2}{*}{toxic (44); total (190)} \\
BAD4 & \\
\hdashline
Stormfront & hate (239); total (478) \\
\hdashline
US-election & hateful (31); total (300) \\
\hdashline
GHC & calls for violence (24); total (5510) \\
\hdashline
CAD & counter-speech (66); total (5307) \\
\hdashline
Misogyny & misogynistic (73); total (657) \\
\hdashline
CMSB & sexist (181); total (2363) \\
\hline
\end{tabular}
    \caption{Number of label occurrences in downstream tasks test sets.}
    \label{tab:downstream_details}
\end{table}

\subsection{Dataset Statistics}

Our benchmark consists of English classification datasets that contain tasks related to problematic content detection. Each label from each dataset is treated as a separate task and we only used tasks with more than 100 positive examples in their training sets. Table  \ref{tab:upstream_details} and \ref{tab:downstream_details} show dataset breakdown along with the number of positive samples per task for downstream and upstream tasks, respectively.



\captionsetup[table]{skip=5pt}
\renewcommand{\arraystretch}{1.3}
\begin{table}[H]
\centering
\begin{tabular}{|l|}
\hline
\begin{tabular}[c]{@{}l@{}}\textbf{Source}: Twitter (6);  Reddit (2);  Wikipedia (2); Gab (1) ; Stormfront (1);  \\ Chat dialogue (4);  Synthetically generated (2);  Civil Comments (1).\end{tabular} \\ \hline
\end{tabular}
\caption{Number of datasets by source. }
\label{tab:source}
\end{table}

\subsection{Model Implementation Details}
Following \citet{jin-etal-2021-learn-continually}, we used a weight generator network with a two-layer MLP model for each adapter and classification head with a hidden size of 32. The weight generator takes a task representation as input and generates the weights for all adapters. 

The BiHNet uses
two forms of task representation; long task representation and short task representation as input to generate the model weights. Long task representation is computed by averaging the embedding of all text samples in the training split of a dataset while short task embedding, which is designed to help the model in few-shot settings, is computed by averaging embeddings of 64 texts sampled from the training set. For both long and short task representations,  we used sentence-transformers \citep{reimers2019sentence} \footnote{\url{https://huggingface.co/sentence-transformers/paraphrase-xlm-r-multilingual-v1}}  with mean pooling. The final model weights are calculated as the sum of weights generated using long and short task representations.


To implement BART-Adapter models, we added an adapter between each layer of BART transformers as an autoencoder model with input and output layer with a size equal to embedding dimensions with one hidden size of 256 in the middle. 
In the multi-task setting, we used hard parameter sharing, sharing only the adapter parameters, and  used a separate classification heads for each task. 
Due to the paucity of problematic content online  most of the datasets in this benchmark are heavily sparse. This sparsity poses challenges to the optimization process. To address this, we used a weighted random sampler ensuring each batch consists of at least 30\% positive samples. 
 
For all experiments, we used a batch size of 32 and trained the models for at most 100 epochs. To prevent the model from overfitting, we used early stopping with a patience of three and chose the best model based on the $F_1$ score. For BiHNet-Reg and BiHNet-EWC, we set the regularization coefficient to 0.01.

\subsubsection{Few-shot training and Evaluation}

We conducted few-shot training for 800 epochs with a batch size of 8 for 8-shot experiments. For 16-shot and 32-shot experiments, we used a batch size of 16. Since the total number of training samples is less than 64 in our downstream few-shot adaptations, we only use the long task representation for BiHNet models.
For BART-Adapter-Multitask, we initialize a new classification head for each downstream task. However, for the BART-Adapter-Vanilla model we keep the existing classification head.

\begin{figure}[H]
    \centering
    \includegraphics[width=0.9\textwidth, height=65mm]{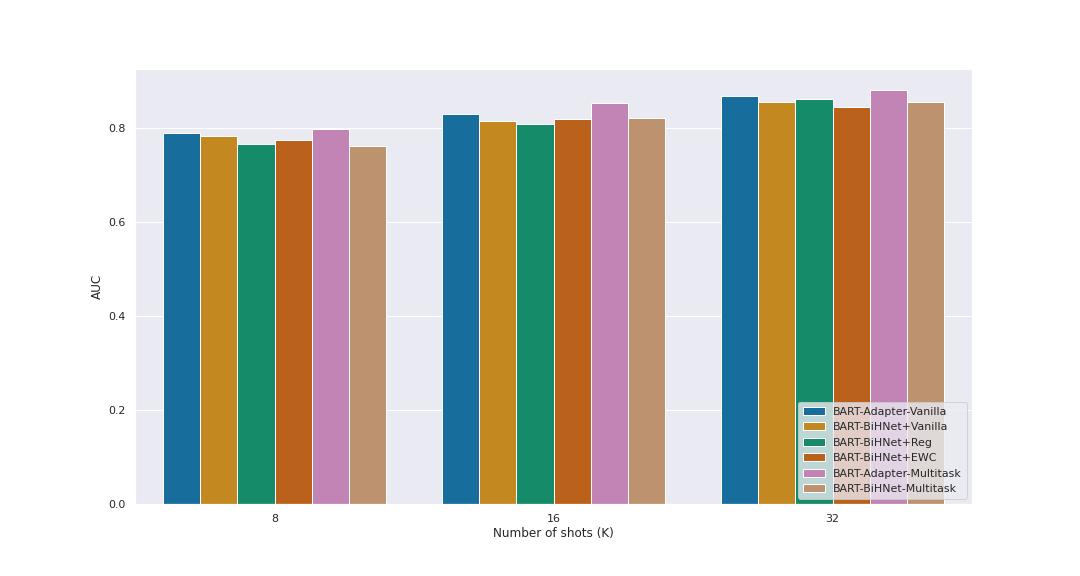}
    \caption{Few-shot performance (AUC) based on number of shots (K)}
    \label{fig:auc_kshot}
\end{figure}

We performed a sensitivity analysis on the number of shots to examine how it affects our models. Specifically, we conducted few-shot training using 8, 16, and 32 shots. You can find the corresponding results in Figure \ref{fig:auc_kshot}. Our results show a consistent pattern; all models improve as the number of shots increases and the order between models stays the same. Interestingly, there is only one exception. BiHNet+Reg outperforms BiHNet+Vanilla with more shots. We leave further investigation of this effect is left for future work.

\label{sec:few-shot-sup}

\subsection{Additional Experiment: Random Upstream Order}
\label{sec:random-exp}
\begin{figure}[htbp]
    \centering
    \includegraphics[width=\textwidth]{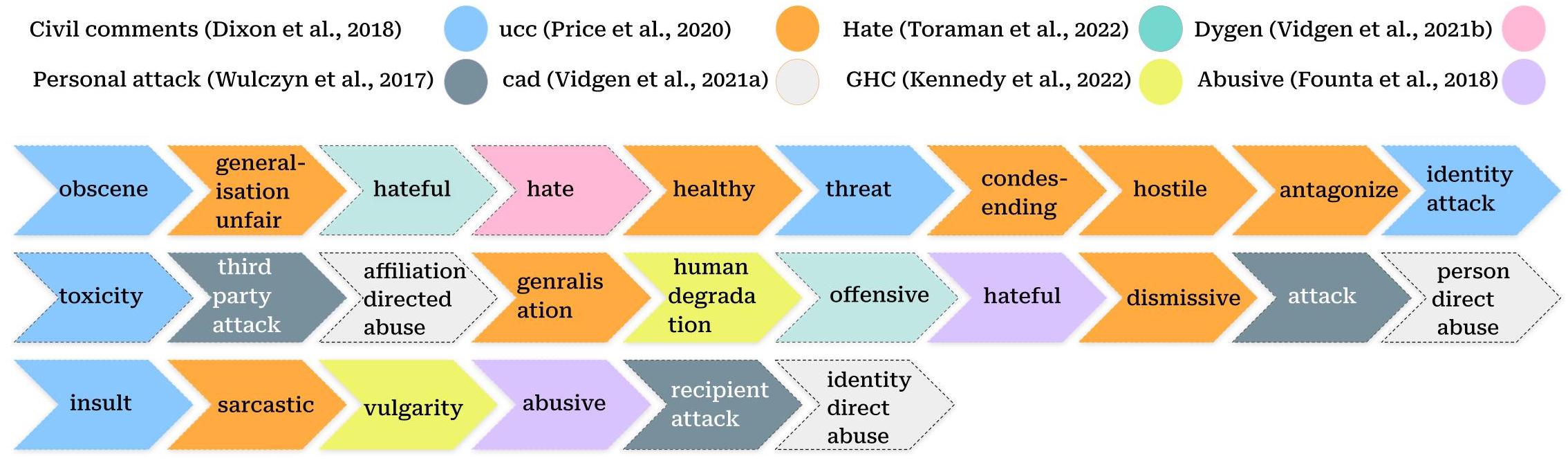}
    \caption{Sequence of upstream tasks used in our benchmark. See section \ref{datasets} for more details.}
    \label{fig:tasks}
\end{figure}

 \pr{To show the efficacy of our proposed continual learning approach in adapting to any scenario, we randomly ordered the upstream tasks.} Figure \ref{fig:tasks} shows upstream task sequence used in our experiments. \pr{Note that, we kept the dataset splits (i.e. train/dev/test) consistent with chronological experiment. This approach ensures that our comparison remains fair and valid, allowing for a meaningful assessment of the model's performance under the altered evaluation conditions.}
 
\subsubsection{Results}
\pr{Overall, we observe similar performance patterns among the different algorithms, but the differences in performance are now less pronounced (Table \ref{tab:results}). Below we discuss the results in detail;}

\noindent\textbf{Single Task Baseline:}
Similar to our initial experiment, to determine the learning capabilities of each model architecture for different tasks \ref{sec:models}, we \pr{finetune} a classifier from each architecture on each task. The average performance of BART, BART-Adapter, and BART-HNet is presented in the first three rows of table \ref{tab:results}. We see that the performance for learning a single task drops as we reduce the number of trainable parameters. 
\captionsetup[table]{skip=5pt}
\renewcommand{\arraystretch}{1.3} 
\begin{table}[ht]
\begin{tabular}{cccccc}
                                            & \multicolumn{2}{c}{Upstream} & Downstream  &                &          \\ \cline{2-6}
                                            & Final      & Instant     & Few-shot  & $\Delta$ Instant & $\Delta$ Few-shot \\ \hline
\multicolumn{1}{l|}{BART-Single}   
&   ------        &     0.8846          &      0.8041       &               ------ &    ------      \\
\multicolumn{1}{l|}{BART-Adapter-Single}        
&   ------        &   0.8791            &     0.8068        &        ------        &      ------    \\
\multicolumn{1}{l|}{BART-BiHNet-Single}        
&   ------       &     0.8702          &      0.7862       &        ------        &      ------    \\
\hline
\multicolumn{1}{l|}{BART-Adapter-Vanilla}  
&   0.6784      &      0.8859        &      0.8321         &   +0.0013          &   +0.0253                      \\
\multicolumn{1}{l|}{BART-BiHNet+Vanilla}        
&   0.7115     &       0.8838    &     0.8146        &    +0.0136            &   +0.0284       \\

\cline{2-6}
\multicolumn{1}{l|}{BART-BiHNet+Reg}      
&   0.7859      &      0.8846        &       0.8087        &      +0.0144        &     +0.0225            \\
\multicolumn{1}{l|}{BART-BiHNet+EWC}        
&   0.6571       &      0.8863         &        0.8190     &   +0.0161             &   +0.0328       \\

\hline
\multicolumn{1}{l|}{BART-Adapter-Multitask} 
& 0.8752 &    ------  &    0.8531           &             ------ &      +0.0463            \\
\multicolumn{1}{l|}{BART-BiHNet+Multitask}        
&    0.8321          &    ------           &    0.8215         &        ------        &   +0.0353       \\
\end{tabular}
\caption{Upstream and Downstream Results. }
\label{tab:results}
\end{table}


\noindent\textbf{Multitask Upperbound:}
The final and few-shot evaluation results for multitask models are displayed in the last two rows of table \ref{tab:results}. It is important to note that these models, having been exposed to all tasks simultaneously, do not have an instant performance metric defined for them.

\noindent\textbf{Does the collection of problematic content tasks help with learning new upstream tasks?}
 To address this inquiry, we can assess the immediate performance of a continual learning (CL) model when applied to $[T_1^u, T_2^u, ..., T_i^u]$ and compare it to a pretrained model fine-tuned exclusively on $T_i^u$. Our results ($\Delta$ Instance) show evidence of slight positive transfer, however, the magnitude of this transfer is negligible.

\noindent\textbf{Does continual learning improve knowledge retention?}
The final AUC values, as shown in the first column of Table \ref{tab:results}, indicate the models' ability to retain knowledge from a sequence of tasks at the end of training. Our results suggest that continual learning (BiHNet+Reg) outperforms naive training (BiHNet+Vanilla) by at least 0.07 in AUC, indicating its potential to mitigate catastrophic forgetting. However, BiHNet+Reg falls 0.04 short of the multitask counterpart. Further investigation is needed to understand this difference.

\noindent\textbf{Does upstream learning help generalize new manifestations of problematic content?}
Comparing the single-task baselines with continual and multitask learning, our results demonstrate a noteworthy improvement in models' generalization ability due to upstream training.

\subsection{Sensitivity Analysis}
\subsubsection{Influence of Task Order in Chronological Experiments}
\begin{figure}[htbp]
    \centering
    \includegraphics[width=\textwidth]{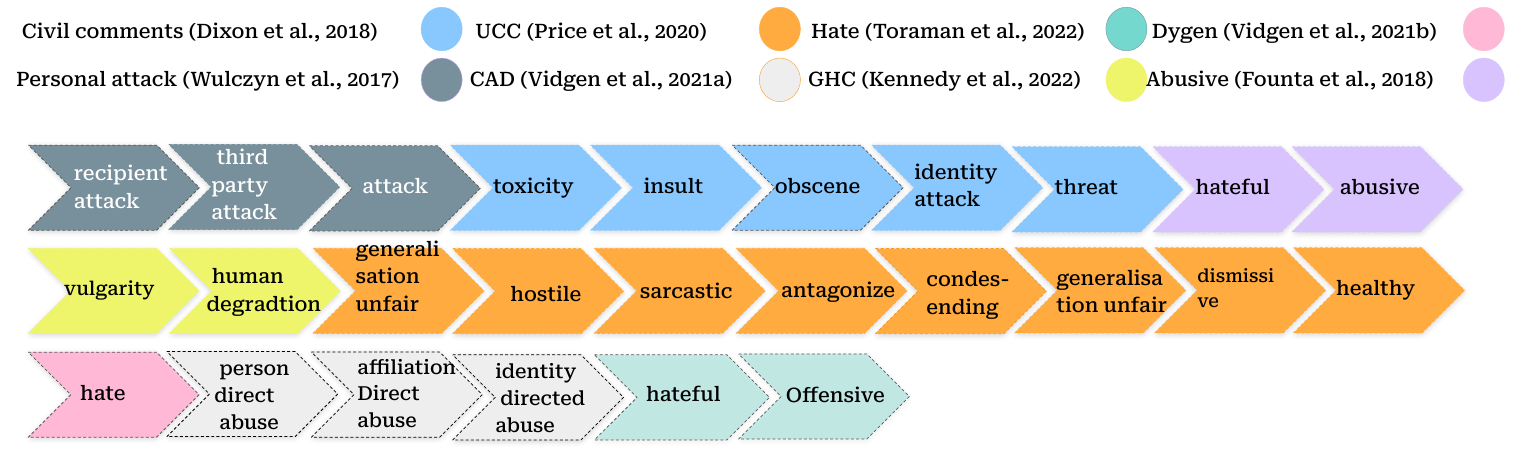}
    \caption{Shuffled sequence of tasks for the chronological experiment.}
    \label{fig:shuffled_chrono_task_order}
\end{figure}

\pr{
In our chronological experiment, we initially assigned tasks within each dataset in a random order, as we lacked information regarding their precedence. To gauge the potential influence of the selected task sequence on our results, we train all model variations again but use an alternative random task order reshuffling while maintaining the dataset order intact. The sequence of upstream tasks in this experiment is illustrated in figure} \ref{fig:shuffled_chrono_task_order}.

\pr{Our results reflect a similar pattern as the initial experiment (Table \ref{tab:combined-shuffled-chrono}) Specifically, the few-shot AUC of BiHNet-Reg improves by nearly 2\% compared to BiHNet Vanilla, falling only 1.2\% short of BiHnet-Adapter-Multitask. In terms of the final AUC, once again, BiHnet-Reg outperforms all sequential fine-tuning variations, and the instant AUC of all models falls within a close range. Overall, this experiment suggests that our proposed approach is robust to task perturbations within datasets. In other words, while the order of tasks within a dataset affects the resulting model's performance, the order of performance among different algorithms remains consistent.}

\begin{table}[H]
    \centering
    \begin{tabular}{lccc}
    \toprule
    model & Fewshot-AUC &  Final-AUC & Instant-AUC \\
    \midrule
    \pr{BART-Adapter-Vanilla} & \pr{0.756767}  & \pr{0.764753} & \pr{0.884386}  \\
    \pr{BART-BiHNet-Vanilla} & \pr{0.786536}  & \pr{0.759363} & \pr{0.881459}  \\
    \pr{BART-BiHNet-Reg} & \pr{0.804270}  & \pr{0.796270} & \pr{0.882993}  \\
    \pr{BART-BiHNet-EWC} & \pr{0.770214}  & \pr{0.751294} & \pr{0.878256}  \\
    \pr{BART-Adapter-Multitask} & \pr{0.816277}  & \pr{0.872739} & \pr{-}  \\
    \pr{BART-BiHNet-Multitask} & \pr{0.795745}  & \pr{0.833905} & \pr{-}   \\
    \bottomrule
    \end{tabular}
    \caption{Combined performance for the models on the chronological experiment}
    \label{tab:combined-shuffled-chrono}
\end{table}

\subsubsection{Effect of Upstream Training on Downstream Few-shot Performance}
\begin{figure}[htbp]
    \centering
    \includegraphics[width=\textwidth]{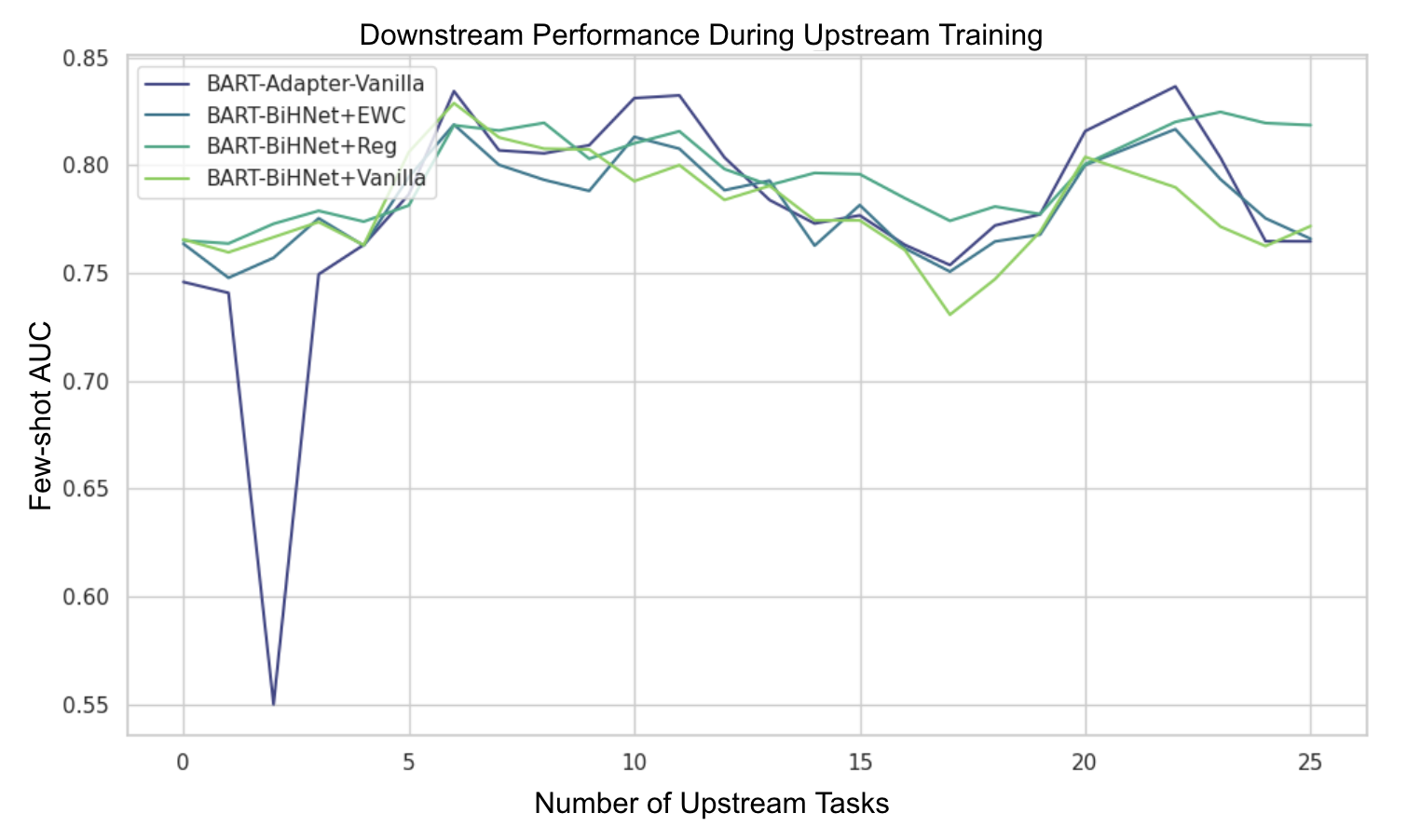}
    \caption{few-shot performance of models at various stages of upstream training.}
    \label{fig:downstream-during-upstream}
\end{figure}

\pr{In this experiment, we ask whether exposing the model to more tasks during the upstream training improves the generalization ability of the model to downstream tasks.To investigate this, we take the model after it has completed training on the \textit{i}th upstream task and proceed to adapt it to all downstream tasks. Figure \ref{fig:downstream-during-upstream} illustrates the average AUC of the models across all downstream tasks as the upstream training advances. As shown in figure \ref{fig:downstream-during-upstream} all algorithms benefit from upstream training. It is important to note that the continual learning algorithms show a more robust behavior in this improvement compared to naive baselines. For example, we see a notable drop in downstream performance of the adapter-vanilla model after task 2 but BiHNet variations remain stable throughout.}

\subsection{Detailed Results}

Table \ref{tab:results-f1} provides $F_1$ scores for models trained on upstream tasks \pr{with random task order}. Note that as we discussed in the paper due to the scarcity of positive labels in all datasets used in our benchmark, AUC is a better suited measure of performance than $F_1$.

\captionsetup[table]{skip=5pt}
\renewcommand{\arraystretch}{1.3} 
\begin{table}[H]
\begin{tabular}{cccccc}
& \multicolumn{2}{c}{Upstream} & Downstream  &                &          \\ \cline{2-6}
& Final      & Instant     & Few-shot  & $\Delta$ Instnat & $\Delta$ Few-shot \\ \hline
\multicolumn{1}{l|}{BART-Single}   
&   ------       &         0.3495      &    0.3398         &        ------        &      ------    \\
\multicolumn{1}{l|}{BART-Adapter-Single}        
&   ------       &       0.4015        &     0.3372        &        ------        &      ------    \\
\multicolumn{1}{l|}{BART-BiHNet-Single}        
&   ------       &      0.4022         &       0.3052      &        ------        &      ------    \\
\hline
\multicolumn{1}{l|}{BART-Adapter-Vanilla}  
&      0.1916    &        0.4190       &      0.3113       &    +0.0175          &      - 0.0259     \\
\multicolumn{1}{l|}{BART-BiHNet+Vanilla}        
&     0.2081     &         0.4180      &     0.2876       &          +0.0158    &       - 0.0176    \\

\cline{2-6}
\multicolumn{1}{l|}{BART-BiHNet+Reg}      
&   0.3058       &        0.4041       &     0.2903       &        +0.0019       &      - 0.0149     \\
\multicolumn{1}{l|}{BART-BiHNet+EWC}        
&     0.1904     &       0.4211        &      0.2856      &        +0.0189         &        - 0.0196   \\

\hline
\multicolumn{1}{l|}{BART-Adapter-Multitask} 
&      0.3975       &    ------           &     0.3294        &        ------        &     - 0.0078      \\
\multicolumn{1}{l|}{BART-BiHNet+Multitask}        
&     0.3183        &    ------           &      0.2722       &        ------        &       - 0.033    \\
\end{tabular}
\caption{F1 scores for both upstream and downstream. }
\label{tab:results-f1}
\end{table}

\subsubsection{Upstream Detailed Results}
\label{sec:sup-result-up-detail}
\begin{small}
\setlength{\tabcolsep}{2pt}
\begin{longtable}{llllrlrl}
\toprule
order & dataset & task & model & final-f1 & instant-f1 & final-auc & instant-auc \\
\midrule
0 & personal-attack & a & BART-Adapter-Vanilla & 0.305006 & 0.750957 & 0.540933 & 0.962732 \\
0 & personal-attack & a & BART-BiHNet+Vanilla & 0.265491 & 0.749760 & 0.727950 & 0.957005 \\
0 & personal-attack & a & BART-BiHNet+Reg & 0.743326 & 0.751853 & 0.956610 & 0.959739 \\
0 & personal-attack & a & BART-BiHNet+EWC & 0.295845 & 0.737895 & 0.896827 & 0.954632 \\
- & personal-attack & a & BART-Adapter-Multitask & 0.703736 & - & 0.957941 & - \\
- & personal-attack & a & BART-BiHNet-Multitask & 0.747288 & - & 0.954593 & - \\
\hdashline
1 & personal-attack & tpa & BART-Adapter-Vanilla & 0.062567 & 0.321267 & 0.461003 & 0.948346 \\
\end{longtable}%
\end{small}

\subsubsection{Downstream Detailed Results}
\label{sec:sup-result-down-detail}
\begin{small}
\setlength{\tabcolsep}{2pt}
\begin{longtable}{lllcc}
\toprule
dataset & task & model & few-shot-auc & few-shot-f1 \\
\midrule
BAD2 & - & BART-Adapter-Vanilla & 0.626491 & 0.475584 \\
BAD2 & - & BART-BiHNet+Vanilla & 0.591835 & 0.442589 \\
BAD2 & - & BART-BiHNet+Reg & 0.627312 & 0.469799 \\
BAD2 & - & BART-BiHNet+EWC & 0.624396 & 0.483940 \\
BAD2 & - & BART-Adapter-Multitask & 0.643871 & 0.492441 \\
BAD2 & - & BART-BiHNet-Multitask & 0.661902 & 0.482916 \\
\hdashline

BAD4 & - & BART-Adapter-Vanilla & 0.590429 & 0.335484 \\
\end{longtable}
\end{small}

\begin{small}
\setlength{\tabcolsep}{2pt}
\begin{longtable}{lllcc}
\toprule
dataset & task & model & AUC & F1 \\
\midrule
BAD2 & - & BART-Single & 0.635964 & 0.490090 \\
BAD2 & - & BART-Adapter-Single & 0.654797 & 0.483221 \\
BAD2 & - & BART-BiHNet-Single & 0.620018 & 0.467909 \\
BAD2 & - & BART-Adapter-Vanilla & 0.678801 & 0.475962 \\
BAD2 & - & BART-BiHNet+Vanilla & 0.582984 & 0.435165 \\
BAD2 & - & BART-BiHNet+Reg & 0.660194 & 0.491484 \\
BAD2 & - & BART-BiHNet+EWC & 0.633916 & 0.470588 \\
\end{longtable}
\end{small}

\end{document}